\documentclass[conference,10pt,a4paper]{IEEEtran}
\IEEEoverridecommandlockouts
\usepackage{cite}
\usepackage{amsmath,amssymb,amsfonts}
\usepackage{algorithmic}
\usepackage{graphicx}
\usepackage{rotating}
\usepackage[leftcaption]{sidecap}
\usepackage{textcomp}
\usepackage[table,dvipsnames]{xcolor}
\usepackage{etoolbox}

\graphicspath{{figures/}}

\usepackage{listings}
\definecolor{codegray}{rgb}{0.5,0.5,0.5}

\definecolor{maroon}{rgb}{0.5,0,0}
\definecolor{darkgreen}{rgb}{0,0.5,0}

\lstdefinelanguage{XML}
{
  basicstyle=\ttfamily\tiny,
  morestring=[s]{"}{"},
  morecomment=[s]{?}{?},
  morecomment=[s]{!--}{--},
  commentstyle=\color{darkgreen},
  moredelim=[s][\color{black}]{>}{<},
  moredelim=[s][\color{red}]{\ }{=},
  stringstyle=\color{blue},
  identifierstyle=\color{maroon}
}

\ifCLASSOPTIONcompsoc
\usepackage[caption=false,font=normalsize,labelfont=sf,textfont=sf]{subfig}
\else
\usepackage[caption=false,font=footnotesize]{subfig}
\fi

\def\BibTeX{{\rm B\kern-.05em{\sc i\kern-.025em b}\kern-.08em
    T\kern-.1667em\lower.7ex\hbox{E}\kern-.125emX}}
    
\usepackage{algorithmic}

\usepackage[linesnumbered,ruled,noend]{algorithm2e}

\usepackage{tikz}

\usepackage{verbatim}

\usepackage{dirtytalk}

\begin{document}

\title{A Multi-threading Kernel for Enabling Neuromorphic Edge Applications}

\author{\IEEEauthorblockN{
Lars Niedermeier\IEEEauthorrefmark{1}\IEEEauthorrefmark{3}, 
Vyom Shah\IEEEauthorrefmark{2}, and
Jeffrey L. Krichmar\IEEEauthorrefmark{2}\IEEEauthorrefmark{3}}
\IEEEauthorblockA{\IEEEauthorrefmark{1}\textit{Niedermeier Consulting, Zurich, ZH, Switzerland}}
\IEEEauthorblockA{\IEEEauthorrefmark{2}\textit{Department of Computer Science}, \textit{University of California, Irvine, CA, USA}}
\IEEEauthorblockA{\IEEEauthorrefmark{3}\textit{Department of Cognitive Sciences}, \textit{University of California, Irvine, CA, USA}}
\IEEEauthorblockA{Correspondence Email: lars@niedermeier-consulting.ch}  
}

\maketitle

\begin{abstract}
Spiking Neural Networks (SNNs) have sparse, event-driven processing that can leverage neuromorphic applications.  In this work, we introduce a multi-threading kernel that enables neuromorphic applications running at the edge, meaning they process sensory input directly and without any up-link to or dependency on a cloud service. The kernel shows speed-up gains over single thread processing by a factor of four on moderately sized SNNs and $1.7X$ on a Synfire network. Furthermore, it load-balances all cores available on multi-core processors, such as ARM, which run today's mobile devices and is up to 70\% more energy efficient compared to statical core assignment. The present work can enable the development of edge applications that have low Size, Weight, and Power (SWaP), and can prototype the integration of neuromorphic chips.

\end{abstract}


\section{Introduction}
\label{sec:intoduction} 

Spiking Neural Networks (SNNs) mimic natural nervous systems with elements that replicate the sparse, all-or-none neural activity. This makes them a good fit to take advantage of low Size, Weight, and Power (SWaP) computing systems. 
The downside of SNNs are the higher computational costs for the numerical solution of the differential equations that determine the neuron's state.
Software packages such as CARLsim have evolved over the years to provide a mature framework for computational neuroscientists, embedded systems engineers, and roboticists \cite{Niedermeier2022}. 

To measure, the potential of SNNs for neuromorphic edge applications, methods are needed to quantitatively measure performance. Recently, a multilayered recurrent SNN benchmark, called a Synfire chain, was used to measure the energy efficiency of the SpiNNaker2 neuromorphic chip \cite{hoeppner2022}.  
The Synfire network, which consisted of leaky-integrated-and-fire (LIF) spiking neurons, is used to measure the dynamic voltage and frequency scaling (DVFS) developed for SpiNNaker \cite{hoppner2017}. 

\begin{figure}[h]
\centering
\subfloat[Synfire network as CARLsim kernel benchmark.]{\includegraphics[width=1.0\columnwidth]{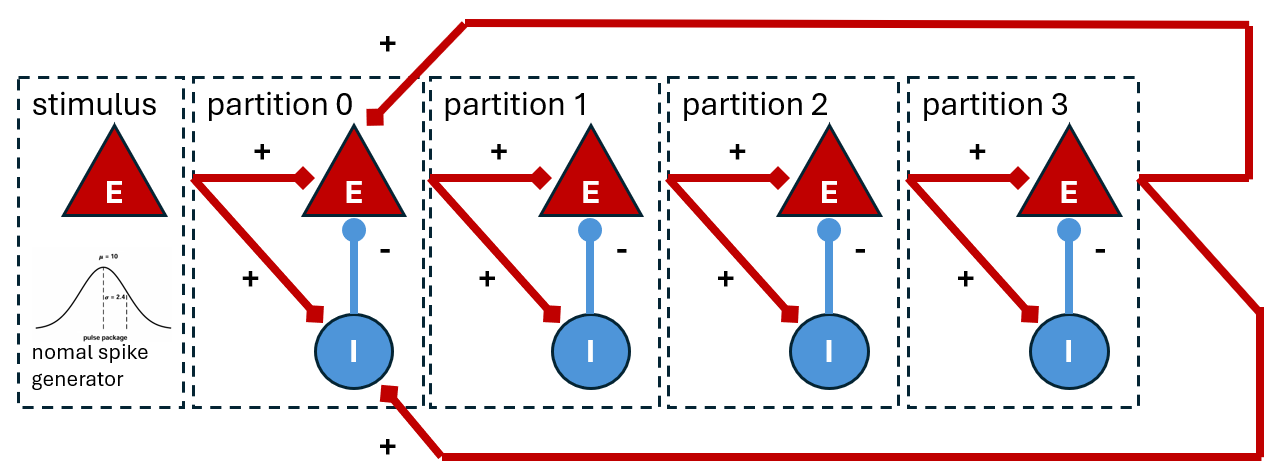}\label{fig:synfire_carlsim}}    
\vfil
\subfloat[Spikewave propagation with correlated inhibition.]{\includegraphics[width=0.85\columnwidth]{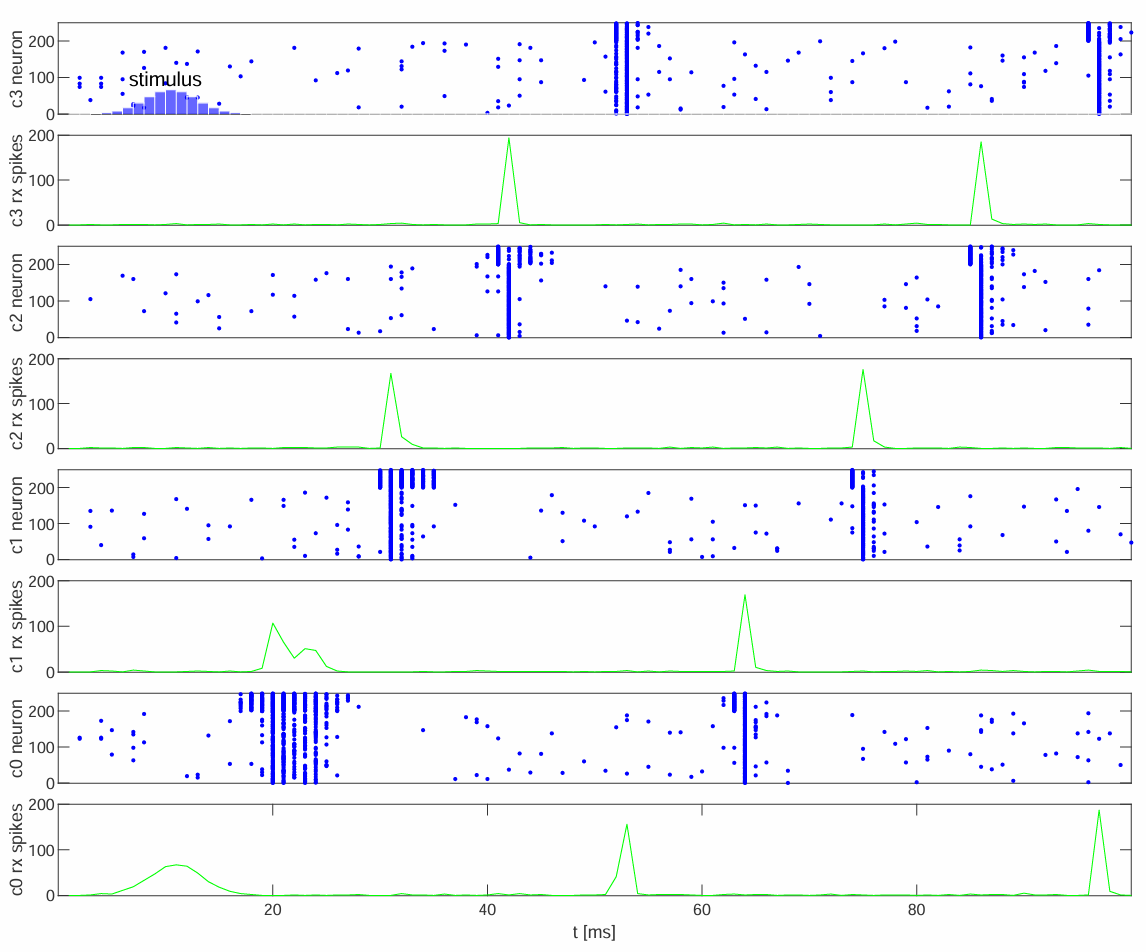}\label{fig:synfire_spikemon}}   
\vfil
\subfloat[PThreads kernel (CARLsim 4).]{\includegraphics[width=0.5\columnwidth]{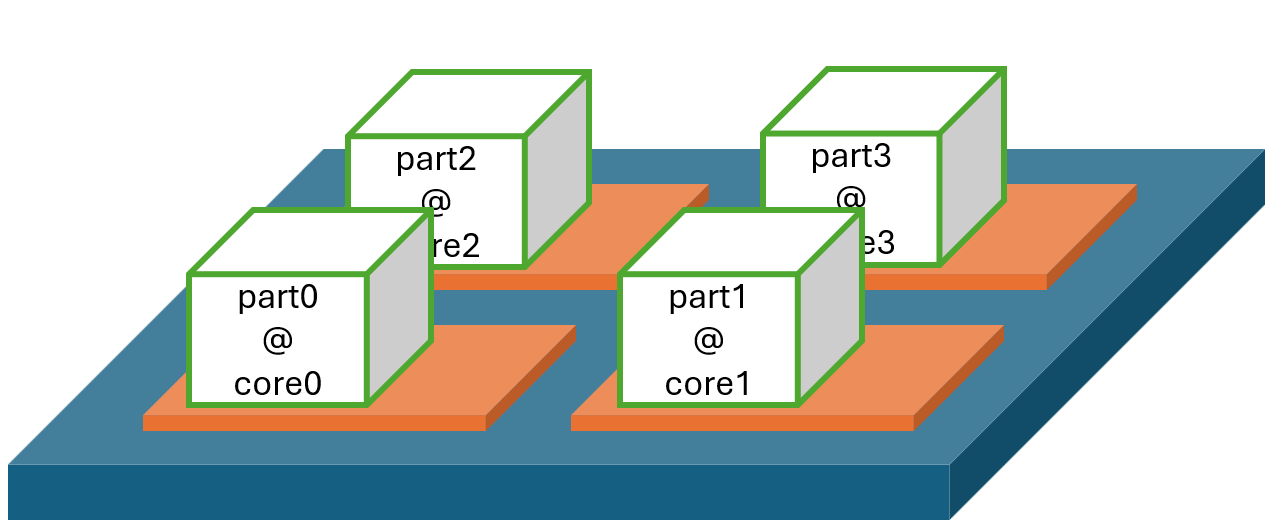}\label{fig:synfire_pthreads_kernel}}    
\subfloat[New multi-threading kernel.]{\includegraphics[width=0.5\columnwidth]{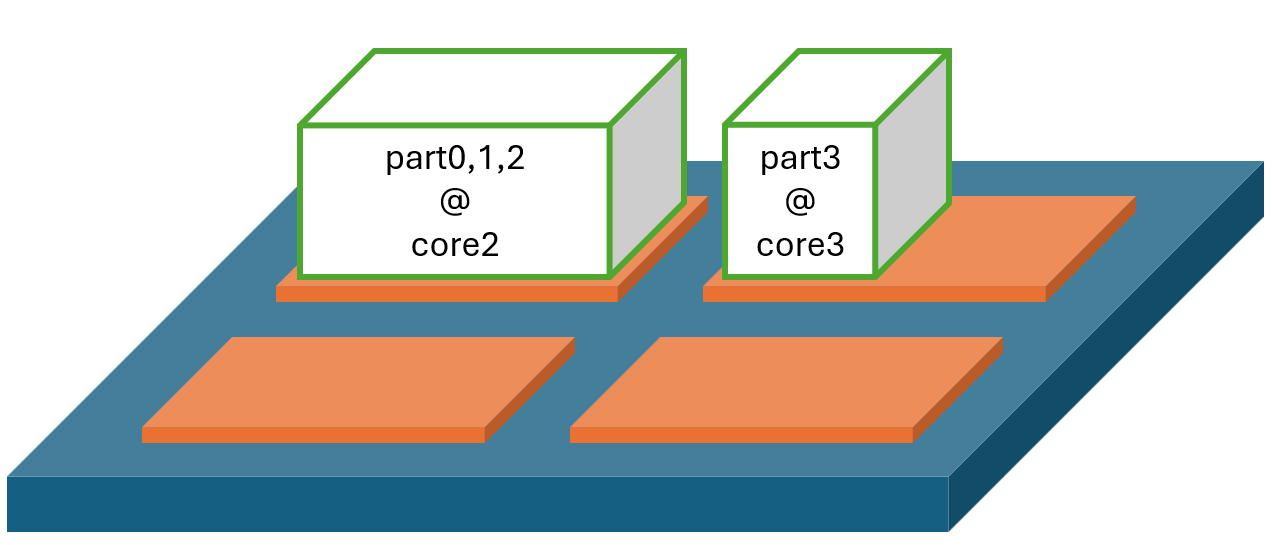}\label{fig:synfire_omp_kernel}}    
\caption{The CARLsim \emph{Synfire} network with normal spike generator for the stimulus.  
(a) Izhikevich neurons in four partitions recurrently linked.
(b) Neural activity visualized in CARLsim spike monitor.
(c) Paritions with fixed core affinity in pthreads kernel of CARLsim4.  
(d) The new kernel is more energy efficient by assigning cores dynamically to free up SoC compute resources.
}
\label{fig:synfire}
\end{figure}

In the present work, we investigate the Synfire benchmark with CARLsim and the Izhikevich neuron model \cite{Izhikevich2003, Izhikevich2004}. Compared to LIF neurons, the Izhikevich neuron has a wider range of dynamics and can mimic a variety of neuron types found in the brain.   
Fig. \ref{fig:synfire} presents the Synfire network developed to benchmark CARLsim. The excitatory groups \emph{E} consist of regular spiking (RS) neurons found as pyramidal cells in the cortex, 
the inhibitory groups \emph{I} of fast spiking (FS) interneurons. 
The network is segmented in four partitions that are assigned to a dedicated core. We follow \cite{hoppner2019} in connecting the last segment $partition3$ to the first $paritition0$
in contrast to the original referenced Synfire network is a strict feed-forward-inhibition (FFI) network \cite{kremkow2010}. 

Depending on the network size, efficient operation may require parallel processing on GPUs, multi-core CPUs, or neuromorphic hardware.
For instance, \emph{small} networks up to a few hundred neurons run most efficiently 
in a single thread on a CPU. 
The efficient simulation of \emph{large-scale} networks with millions of neurons, 
as typically used in computational neuroscience \cite{Kopsick2022}, are the core feature of CARLsim as its PThreads based kernel scales over multiple GPUs \cite{ChouHirak2018}. In the case of \emph{mid-size} networks (e.g., $10^3$ neurons), which are often used for edge applications, the overhead for PThreads has poor performance compared to an execution on a single thread. 

In the present work, we introduce a multi-threading kernel that scales efficiently over the available cores of modern CPUs used in SoCs such as ARM Cortex-76 in Raspberry Pi 5. This addresses the performance limitation of PThreads in CARLsim and other SNN simulators.
All code and models are open-source and available on \cite{CARLsim6Repository}.  
The main contributions of this work are:

\begin{enumerate}
\item \textbf{\textit{Multi-threading for SNNs.}} A multi-threading kernel that scales independently of the partitioning of the network. 
\item \textbf{\textit{Dynamic load balancing.}} A load-balancing algorithm that dynamically allocates computation to cores and avoids synchronization bottlenecks of idle threads.  
\item \textbf{\textit{Performance monitoring.}} An SNN performance monitor with ms precision. 
\item \textbf{\textit{Synthetic load network.}} A synthetic load network named \emph{Chainfire} that produces and measures neural and synaptic activity.
\item \textbf{\textit{SNN Benchmarks.}}  Concrete benchmark results on Intel and ARM processors. 
\item \textbf{\textit{Edge processing.}} SNNs that fulfill real-time requirements on off-the shelf mobile processors, without the need of specialized neuromorphic hardware.  
\end{enumerate}

\section{Methods}
\label{sec:methods} 

\subsection{Spiking neuron model}

CARLsim efficiently implements spiking neural networks such as LIF and the Izhikevich neuron model with 4 and 9 parameters \cite{Niedermeier2022}. 
In contrast to \cite{hoeppner2022}, we implemented the Synfire network utilizing Izhikevich 4-parameter model described by the following equations \cite{Izhikevich2003}.
 
\begin{eqnarray} \label{eq:Izhi4}
\dot{v} &{}={}& 0.04v^2 + 5v + 140 - u + I    \\
\dot{u} &{}={}& a(bv - u)  \label{eq:Izhi4u}
\end{eqnarray}
\begin{equation} \label{eq:Izhi4v30}
\text{if}\ v \geq 30 \begin{cases}
v = c  \\
u = u + d
\end{cases}
\end{equation}

The present simulations use Forward Euler for numerical integration.  More complicated neuron models with multiple compartments may require other numerical methods, such as Runge-Kutta, to handle instabilities. The CARLsim kernel is designed to handle these cases.

\subsection{Synfire network architecture}     
\label{subsec:synfire}
 
We utilize the Synfire chain network as a benchmark to measure and compare performance of CARLsim and potentially other neuromorphic chips. 
We follow Höppner et. al. in their approach for SpiNNaker \cite{hoeppner2022} and built a network with the same structure and sizing, see Fig. \ref{fig:synfire}.
The excitatory groups $E$ has 200 regular spiking (RS) neurons $(a=0.02, b=0.2, c=-65, d=8)$, and the inhibitory groups consist of 50 fast spiking (FS) neurons $(a=0.1, b=0.2, c=-65, d=2)$.
The code and configuration is Open Source and available in the CARLsim GitHub repository \cite{CARLsim6Repository}. 
Using the CARLsim multi-thread kernel, we replicated Synfire results for current based (CUBA) and conductances based (COBA) synapses. See Fig. \ref{fig:energy_savings_vs_fixed_core_allocation}b and GitHub repository \cite{niedermeier2026iscas} for the benchmark runs  and Subsection \ref{subsec:synfire}. 
\color{black}

\subsection{Chainfire network architecture}     
\label{subsec:chainfire}

\begin{figure}[h]
\centerline{\includegraphics[width=0.95\columnwidth]{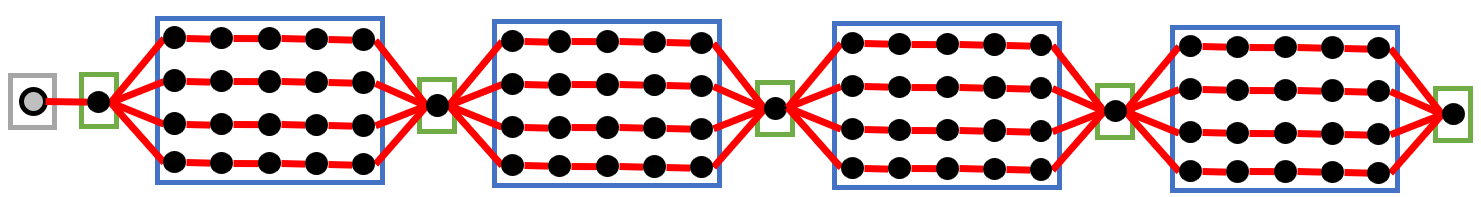}}   
\caption{\emph{Chainfire} SNN for validation, tuning, and scaling of the parallelization. The blue rectangles contain groups of excitatory neurons.
It can induce specific loads on the multi-threading kernel. The green squares depict synchronization neurons that dictate the fan-in and fan-out between neuron groups.}
\label{fig:chainfire}
\end{figure}

In addition to using the Synfire chain SNN for benchmarking, we created a Chainfire network, which could more readily control parameters that affect CPU loads. Fig. \ref{fig:chainfire} shows the synthetic load network \emph{Chainfire} with the minimal amount of synapses that allows to produce and   
measure arbitrary loads, specifically targeted at the compute-intensive status updates of neurons, defined by equations \ref{eq:Izhi4} - \ref{eq:Izhi4v30}.
The network is designed to generate loads similar to the \emph{Synfire} network and therefore has four clusters, that are marked in blue.
This allows validation, tuning, scaling of the parallelization, and provides the maximum of performance improvement possible by this kernel.

The parallel rows are chains of excitatory RS neurons with specific delays and weights (e.g. $d_{exc}$ = 5 ms, $w_{exc}$ = 0.432), 
configured to propagate the spike wave with minimal delay and without noise.
The synchronization neurons, indicated by a green frames, are the also excitatory RS neurons and consolidate the fan-in for predecessor group 
or integrate the fan-out.  

The feed-forward load generator SNN is configurable by the following parameters: 
$N$ total number of neurons per cluster, 
$d$ ms of delay for pre-synaptic to post-synaptic neuron in the parallel chains that are sequential connected, and the 
$span$ in ms that determine the length of the chain, 
e.g.  $N = 100$, $d = 20 \text{ms}$, $span = 100 \text{ms}$ results in the shown network groups with 100 neurons, 
structured in 4 rows (parallel chains) and five columns.  

The fan-in from synchronization neurons are configured, so that a spike activates the input layer (first column) at the same time.  
The spike-wave travels through the group and activates each neuron of the adjacent column.
The output layer (last column) transmits the spikes to the fan-out which has reduced weights, so that the synchronization neurons produce a single spike.
This in turn feeds back to the first column of the Chainfire. 

The frequency of the input spike generator is 1 Hz. The delays enforce a travel time of the spike-wave of around 500 ms (400 ms for the cluster plus inter-neurons, and some processing time around 2 ms for the RS neurons). 
The spike-wave travels through the network without interference to the last synchronization neuron, before the next stimuli is induced.

\subsection{Benchmark processors } 

SNNs running at the edge can take advantage of off-the-shelf multi-core processors. The CARLsim multi-threaded kernel utilizes  available cores of modern energy efficient CPUs, such as the ARM Cortex-76 in Raspberry Pi 5. These processors are typically built in System-on-chips (SoCs), and are used to operate devices at the edge. For a moderate sized SNN to run efficiently at the edge, care must be taken to ensure all cores have fully balanced loads. Algorithm \ref{alg:omp} outlines parallelization used in the multi-threaded kernel. 

\begin{algorithm}[h]
	\scriptsize{
 		\KwIn{simulation time in ms, network partition $net_{id}$, numerical integration steps per ms $S$, core threads $T$}
 		\KwOut{spikes}
        \While(\tcc*[f]{numerical integration}){$step \in S$ } 
        {
            \While(\tcc*[f]{initialize shared}){$group \in partion$ } 
            {                
         		\While(\tcc*[f]{in parallel $T$}){$neuron \in group$} 
                {
                    $I_{sum} = \texttt{CUBA/COBA}$\tcc*[r]{synaptic fan-in}  
                    $v, u = \texttt{EULER/RK4}(dudt, dvdt, a, b, c, d, I_{sum})$\\
                    \If(\tcc*[f]{Izhikevich}){$v > 30$}{$spike = true$;}
                    $runtimeData.recovery[lNId] = u$\tcc*[r]{update}                    
     	          }
            }
        }
	\caption{\footnotesize{Multi-threading kernel for neuron state updates.}}
	\label{alg:omp}
    }
\end{algorithm}

\subsection{Load balancing by dynamic core assignment (DCA)}     
\label{subsec:methods_dca}

One reason SNNs are energy efficient is that their activity is sparse with a firing rate of a few Hz with occasional spike bursts. 
CARLsim utilizes the intrinsic Dynamic Voltage and Frequency Scaling (DVFS) of modern CPUs, which provide advanced energy policies. 
Intel no longer recommends direct manipulation of the P-states to manipulate DVFS and recommends only to use power profiles for the sake of system stability.

The overhead for thread synchronization cannot be neglected.
Consequently, our load-balancer allocates only the minimum cores necessary to fulfill the performance criteria, 
for instance real-time, meaning 1 ms in the SNN corresponds to 1 ms wall clock time.

\section{Results}
\label{sec:results}
We provide benchmark results for the Intel i9 with 8 cores and several ARM Cortex processors with 4 cores. 
All results can be reproduced with the open source of the CARLsim repository.  
Furthermore, we provide supplemental material such as videos and log of the run on GitHub \cite{niedermeier2026iscas}.

\subsection{Performance gain by multi-threading on Chainfire}

\begin{figure*}[ht]
\centering
\subfloat[Chainfire on Intel i9]{\includegraphics[width=0.18\textwidth]{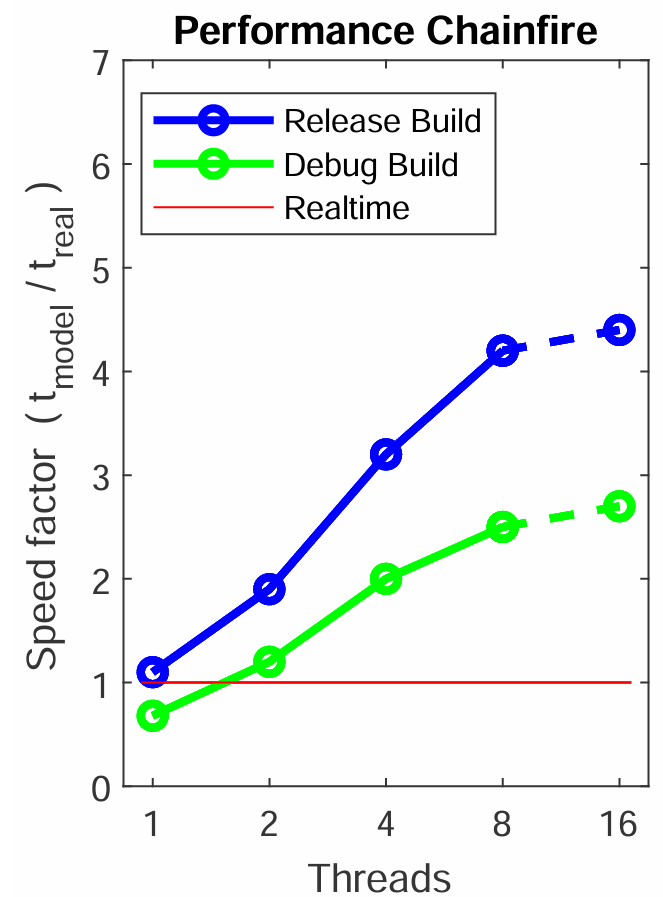}\label{fig:performance_gain_by_mt_chainfire}}    
\subfloat[Synfire on Intel i9 ]{\includegraphics[width=0.315\textwidth]{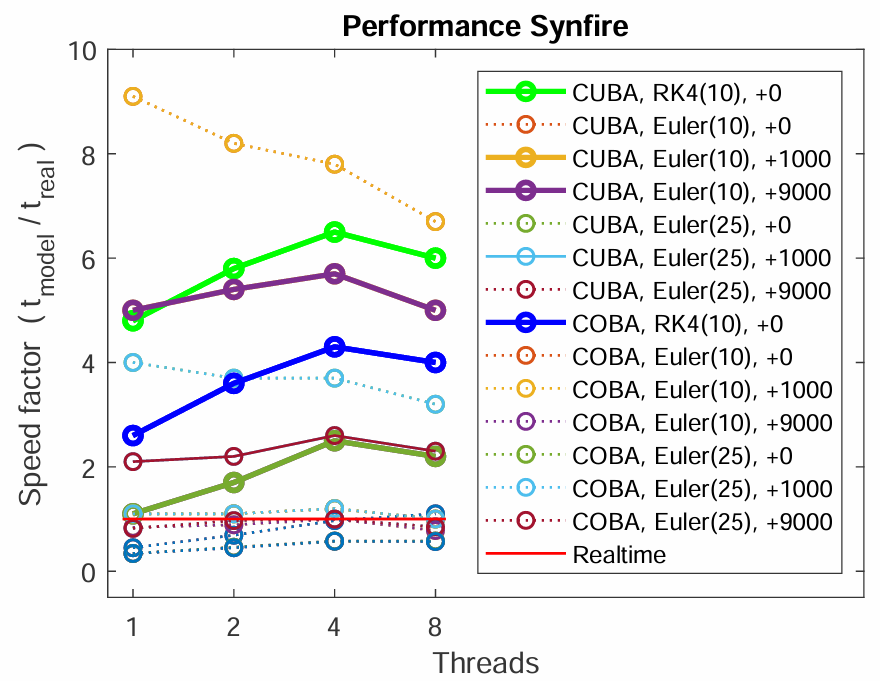}\label{fig:performance_gain_by_mt_synfire}}
\subfloat[Synfire on ARM Cortex]{\includegraphics[width=0.248\textwidth]{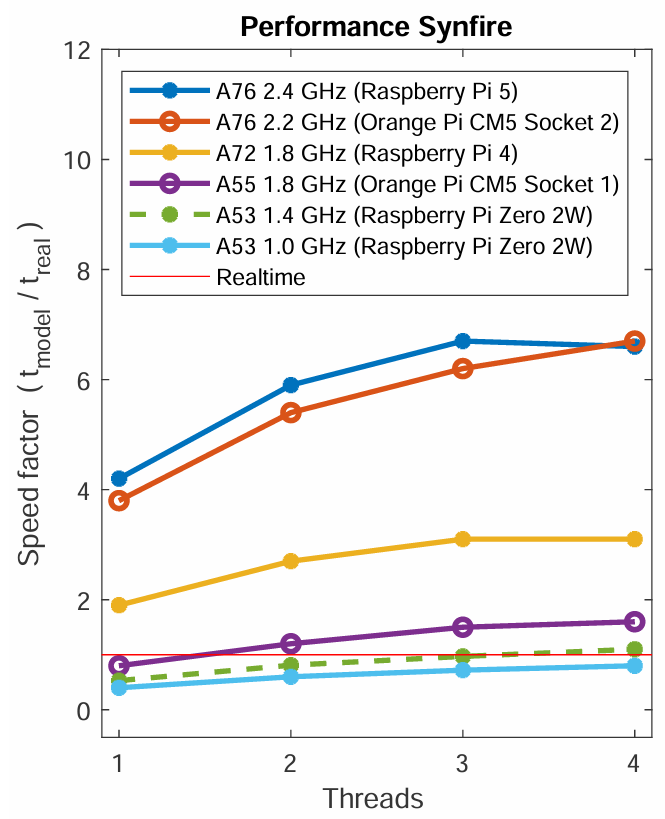}\label{fig:performance_arm}}
\hfil 
\subfloat[DCA]{\includegraphics[width=0.225\textwidth]{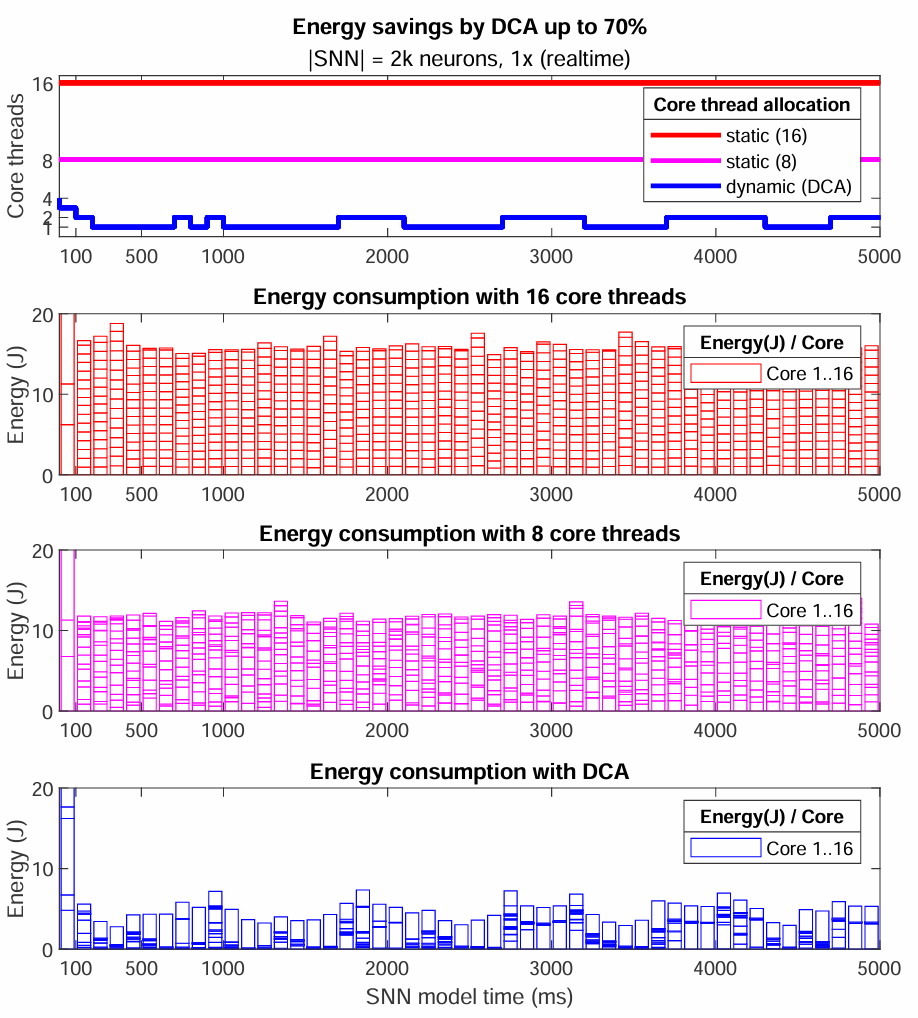}\label{fig:energy_savings_vs_fixed_core_allocation}}
\caption{Performance improvements by multi-threading kernel over allocated cores.
(a) Chainfire (2k neurons) on 11th Gen Intel Core i9-11900K @ 3.5 GHz (8 cores, 16 logical).     
(b) Synfire (1.2k neurons, 77k synapses) on Intel i9.    
(c) Synfire on several ARM Cortex processors.
(d) Dynamic core allocation saves up to 70\% energy as overhead for over-allocated cores thread synchornization is avoided. The performance monitor Intel PCM shows this at ms precision at timeline of the SNN.
}
\label{fig:performance_improvments}
\end{figure*}

We measure Chainfire performance on a release build and a version with debugging enabled.
Fig. \ref{fig:performance_gain_by_mt_chainfire} and Tables \ref{table:chainfire_release} and \ref{table:chainfire_debug}  
present the performance of a Chainfire network with $2000$ neurons run on a Intel i9 desktop processor. 
The simulation model time was 10s. The overall spike count is $20,040$ and applied as a measure to determine that the neural activity is same for all (parallel) runs.   
The $t_{execution}$ ($t_{real}$) is the wall clock time, which is implemented by the C++ standard library (STL) \emph{std::chrono::steady\_clock}.
Performance is measured by a $speed\_factor$, defined as $t_{model} / t_{execution}$.
For example, if the simulation of the SNN with eight threads runs in 2.28s, it has a $speed\_factor = 4.4$.
If the single threaded run takes 8.88s, it a the $speed\_factor = 1.1$. 
The resulting performance gain is then $ 4.4 / 1.1 = 4.0$.

\begin{table}[ht]
\caption{Performance improvements Chainfire (release build)}
\label{table:chainfire_release}
\centering
\begin{tabular}{rrrr}
\hline
\bf Threads	& \bf Execution Time & \bf Speed Factor & \bf Performance Gain \\  
\hline
1	& 8.88 s	& 1.1 x	&   \\  
2	&  5.20 s	& 1.9 x	& 1.7 x \\ 
4	&  3.09 s	& 3.2 x	& 2.9 x \\ 
8	& 2.38 s	& 4.2 x	& 3.8 x \\ 
\bf 16	& \bf 2.28 s	& \bf 4.4 x	& \bf 4.0 x \\ 
32	& 5.87 s	& 1.7 x	& 1.5 x \\ 
\hline
\end{tabular}
\end{table}

\begin{table}[ht]
\caption{Performance improvements Chainfire (debug build)}
\label{table:chainfire_debug}
\centering
\begin{tabular}{rrrr}
\hline
\bf Threads	& \bf Execution Time & \bf Speed Factor & \bf Performance Gain \\  
\hline
1	&  \color{red} 14.82 s	& \color{red} 67.5\%     &   \\   
2	&	8.49 s	& 1.2 x	     & 1.8 x   \\
4	&	4.93 s	& 2.0 x	     & 3.0 x  \\
8	&	3.96 s	& 2.5  x	& 3.7 x   \\
\bf 16	& \bf 3.72 s	& \bf 2.7 x	    & \bf 4.0 x  \\
32	&	5.87 s	& 1.7 x	    & 1.9 x \\
\hline
\end{tabular}
\end{table}

\subsection{Performance gain by multi-threading on Synfire}

Fig. \ref{fig:performance_gain_by_mt_synfire} and Table \ref{table:synfire_release} present 
the performance improvement on the Synfire chain network with 1,200 neurons and 77K synapses
run on a Intel i9 desktop processor.
As expected, the performance gain is lower than on the synthetic SNN as the optimization for the multi-threading kernel
aimed primarily on the compute intense numerical integration of the differential equations for the neuron state. 
On the other hand, the network is rather small and $70\%$ performance gain is actually above expectations. 
Future work will aim to parallelize the synaptic processing.

\begin{table}[ht]
\caption{Performance improvements Synfire (release build)}
\label{table:synfire_release}
\centering
\begin{tabular}{rrrr}
\hline
\bf Threads	& \bf Execution Time & \bf Speed Factor & \bf Performance Gain \\  
\hline
1	& 0.43 s	& 7 x	&   \\  
2	&  0.31 s	& 9.7 x	& 1.4 x \\ 
\bf 4	&  \bf 0.26 s	& \bf 12.0 x	& \bf 1.7 x \\ 
 8	& 0.26 s	& 12.0 x	& 1.7 x \\ 
16	& 0.29 s	& 10.0 x	& 1.4 x \\ 
32	& 0.62 s	& 4.8 x	&  0.7 x\\ 
\hline
\end{tabular}
\end{table}

Fig. \ref{fig:performance_arm} presents the the performance improvements of the same Synfire network on several ARM processors 
that can be used at the edge.
The kernel has a similar structural performance scaling on ARM as on Intel, as long as the cores are used on Intel. 
Because two Intel \emph{logical processors} share a core, the performance degrades, which is indicated by the dashed line in Fig. \ref{fig:performance_gain_by_mt_synfire}.

\subsection{Energy savings by DCA}     
\label{subsec:results_dca}

Fig. \ref{fig:energy_savings_vs_fixed_core_allocation} shows that DCA reduces the core threads as long as the real-time criteria are met (first 250 ms). 
When the system load demands, DCA allocates additional core threads (e.g. at 750 ms and 1800 ms), respectively frees them, when no longer needed (e.g. at 750 ms and 1800 ms).

\section{Conclusion} 
\label{sec:discussion}
The present work introduces a multi-threading kernel for SNNs.  It especially optimizes moderately sized SNNs deployed on the multicore processors used for mobile devices. The kernel supports dynamic core allocation to ensure efficient processing across SoC devices.  We show impressive performance gains on a network architecture, Synfire chain, which is commonly used to test SNNs.  We also introduce a Chainfire network to further evaluate performance.

The synthetic Chainfire network is instrumental to understand the performance relevant internal workings of the former multi-threading kernel.  
It also enabled to validate and quantitative measure the optimization, see Fig. \ref{fig:performance_gain_by_mt_chainfire}. Without it, optimization information is hidden behind noise. There are great tools available for performance profiling. However, to identify bottlenecks, 
it was essential to produce specific loads, for instance the status update only without interference of the synaptic processing. 

With the multi-threading kernel extensions, CARLsim now supports mid-size SNNs on modern multicore processors such as the ARM Cortex family.  Compared to neuromorphic chips such as Intel's Loihi or Brainchip's Akida, which are highly specialized on neural processing 
and usually require a host system to run the base application, 
CARLsim utilizes the existing compute capacity of the mobile processor. Depending on the use case, similar energy efficiency on these specialized neuromorphic chips might be achieved with CARLsim on mobile processors. The new power saving policies in CARLsim may make the intrinsic energy efficiency of SNNs even more attractive for edge applications.  It makes neuromorphic applications possible without additional hardware costs. Furthermore, an SNN executed by software is much more flexible in regards of changes and use case adaptation. 

The present work opens the way for a new generation of neuromorphic applications that can be deployed on millions of mobile devices, such as wearables like the Samsung Watch Ultra 2025
or embedded devices based on SoCs such as the Raspberry Pi 5. 

\bibliographystyle{IEEEtran}
\bibliography{references}

\end{document}